\documentclass[sigconf,natbib=true,anonymous=false]{acmart}
\usepackage{multirow}
\usepackage{subfigure}
\usepackage{makecell}
\usepackage{filecontents}
\usepackage{bbm}
\usepackage{pgfplots}
\usetikzlibrary{patterns}
\usepackage[inline,shortlabels]{enumitem}
\usepackage{natbib}
\usepackage{csquotes}
\usepackage{hyperref}
\usepackage{url}
\usepackage{CJKutf8}
\usepackage{algorithm}
\usepackage{svg}
\usepackage{amsmath}
\usepackage{bm}
\usepackage{booktabs} 
\usepackage{graphics}
\usepackage{graphicx}
\usepackage{array}
\usepackage[english]{babel}
\usepackage[T1]{fontenc}
\usepackage{textcomp}
\usepackage{latexsym}
\usepackage[labelformat=simple]{subcaption}
\usepackage{makecell}

\usepackage{siunitx}  
\usepackage[normalem]{ulem}
\usepackage{array}

\usepackage[most]{tcolorbox}
\usepackage{lipsum} 
\usepackage{float}  

\usepackage{multirow}  
\usepackage{booktabs}  

\usepackage{caption}
\usepackage{subcaption}
\usepackage{graphicx}
\usepackage{titletoc}
\usepackage{amsfonts}
\usepackage{booktabs}
\usepackage{arydshln}
\usepackage{colortbl}
\usepackage{algorithm}
\usepackage[noend]{algpseudocode}
\usepackage{enumitem}
\usepackage{graphicx}
\usepackage{soul}
\usepackage{colortbl,array,xcolor}

\definecolor{citecolor}{HTML}{0051f4}
\definecolor{pink}{HTML}{ed008c}
\usepackage{hyperref}       
\usepackage{url}            
\usepackage{booktabs}       
\usepackage{nicefrac}       
\usepackage{microtype}      
\usepackage{xcolor}         
\usepackage{tcolorbox}
\usepackage{CJKutf8}
\usepackage{multirow}
\usepackage{wrapfig}
\usepackage{capt-of}
\usepackage{graphicx}  
\usepackage{pgfplots}
\pgfplotsset{compat=1.12}
\usepackage{amsmath}
\usepackage{multicol}
\usepackage{color}
\usepackage{mwe}
\usepackage{wrapfig}
\usepackage{colortbl,array}
\usepackage{xspace}
\usepackage{tikz}
\usetikzlibrary{tikzmark}
\usepackage{multirow}  
\begin{document}

\title{Leveraging LLM-Assisted Query Understanding\\ for Live Retrieval-Augmented Generation}


\author{Guanting Dong}
\affiliation{%
  \institution{Renmin University of China}
  \city{Beijing}
  \country{China}}
\email{dongguanting@ruc.edu.cn}

\author{Xiaoxi Li}
\affiliation{%
  \institution{Renmin University of China}
  \city{Beijing}
  \country{China}}
\email{xiaoxi_li@ruc.edu.cn}

\author{Yuyao Zhang}
\affiliation{%
  \institution{Renmin University of China}
  \city{Beijing}
  \country{China}}
\email{2020201710@ruc.edu.cn}

\author{Mengjie Deng}
\affiliation{%
  \institution{Renmin University of China}
  \city{Beijing}
  \country{China}}
\email{dengmengjie_777@163.com}



\begin{abstract}
Real-world live retrieval-augmented generation (RAG) systems face significant challenges when processing user queries that are often noisy, ambiguous, and contain multiple intents. While RAG enhances large language models (LLMs) with external knowledge, current systems typically struggle with such complex inputs, as they are often trained or evaluated on cleaner data. This paper introduces Omni-RAG, a novel framework designed to improve the robustness and effectiveness of RAG systems in live, open-domain settings. Omni-RAG employs LLM-assisted query understanding to preprocess user inputs through three key modules: (1) Deep Query Understanding and Decomposition, which utilizes LLMs with tailored prompts to denoise queries (\textit{e.g.}, correcting spelling errors) and decompose multi-intent queries into structured sub-queries; (2) Intent-Aware Knowledge Retrieval, which performs retrieval for each sub-query from a corpus (\textit{i.e.}, FineWeb using OpenSearch) and aggregates the results; and (3) Reranking and Generation, where a reranker (\textit{i.e.}, BGE) refines document selection before a final response is generated by an LLM (\textit{i.e.}, Falcon-10B) using a chain-of-thought prompt. Omni-RAG aims to bridge the gap between current RAG capabilities and the demands of real-world applications, such as those highlighted by the SIGIR 2025 LiveRAG Challenge, by robustly handling complex and noisy queries.

\end{abstract}

\ccsdesc[500]{Information systems~Retrieval models and ranking}
\keywords{Retrieval-Augmented Generation, Query Understanding, Denoising, Document Ranking}

\maketitle
\title{Omni-RAG: Leveraging LLM-Assisted Query Understanding for Live Retrieval-Augmented Generation}

\section{Introduction}

The rapid advancement of large language models (LLMs)~\cite{DBLP:journals/corr/llmsurvey, llama2, llama3} has led to transformative progress across a wide range of natural language processing tasks~\cite{llm4ir-survey, genir-survey}. Nevertheless, when tackling knowledge-intensive tasks, LLMs still rely solely on their internal knowledge, which often fail short in factual inconsistency and hallucinations~\cite{survey_hallu_llm}. To address these issues, researchers have proposed Retrieval-Augmented Generation (RAG)~\cite{rag-survey}, which incorporates external knowledge sources to assist LLMs in content generation, significantly improving the accuracy and reliability of the outputs.

However, in real-world live RAG applications, user queries are rarely atomic or single-intent. Instead, they often involve multiple intents, complex structures, and various types of noise~\cite{RQ-RAG, PlanRAG}. 
While existing RAG methods perform well on standard benchmarks, they typically select simple and noise-free dataset for fine-tuning or alignment. Consequently, these systems struggle to accurately interpret intent and generate reliable responses when faced with noisy, ambiguous, and multi-intent queries in open-domain settings.

To advance research in this direction, the SIGIR 2025 LiveRAG Challenge introduces the first competition specifically designed to evaluate the real-time problem-solving capabilities of online RAG systems. The challenge provides all teams with a fixed knowledge corpus (FineWeb)~\cite{FineWeb} and a pre-trained language model (Falcon3-10B-Instruct)~\cite{Falcon}, while dynamically generating diverse user queries using a configurable synthetic DataMorgana~\cite{DataMorgana} simulator to mimic live human query interactions. Participating RAG systems must complete the task within a two-hour time limit, requiring efficient handling of complex, noisy, and multi-intent queries. Thus, the core challenge lies in robustly and efficiently understanding the underlying intents and semantic noise in user queries, posing a significant hurdle for building practical live RAG systems.

To bridge this gap, we design Omni-RAG, a framework that leverages the understanding capabilities of LLMs to preprocess user queries—through denoising and intent decomposition—enhancing the robustness of RAG systems in real-world online environments. The framework comprises three key modules:
\begin{itemize}[leftmargin=1em]
\item \textbf{Deep Query Understanding and Decomposition:} Based on LLMs' strengths in language understanding, we apply tailored prompts to guide the model in preprocessing user queries. This includes rewriting noisy inputs and decomposing complex queries with multiple intents into clearer, structured sub-queries.

 \item \textbf{Intent-Aware Knowledge Retrieval:} To retrieve comprehensive and relevant supporting information, we use an OpenSearch system to perform retrieval over the FineWeb corpus for each rewritten and decomposed sub-query. The retrieved documents are then aggregated into a unified corpus that captures a broader semantic context for generation.

\item \textbf{Reranking and Generation:} Before generation, we apply a BGE reranker to reorder candidate documents for all sub-queries, selecting the top-10 most relevant ones. These are then integrated with the rewritten main query into a chain-of-thought prompt, which is fed into the Falcon-10B model to generate the final response.
\end{itemize}
Experimental results show that Omni-RAG framework achieved a \textbf{Rank-2 overall performance in Session 1 of the SIGIR LiveRAG Challenge}. Additionally, we conduct pseudo-labeling experiments on the dry-test set following the official evaluation metrics, further demonstrating the framework's effectiveness in terms of both generation efficiency and factual consistency. 

\begin{figure*}[t]
    \centering
    \includegraphics[width=.95\linewidth]{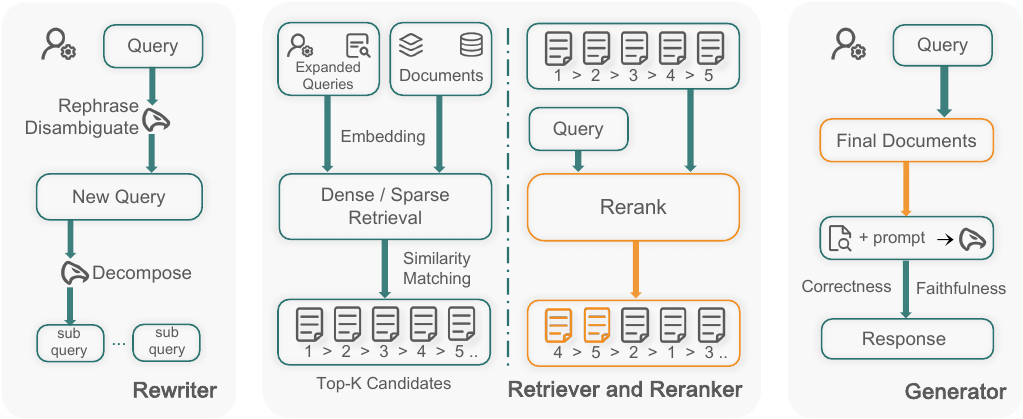}

    \caption{The overall pipeline of our Omni-RAG.}
    \label{fig:main} 
\end{figure*}

\section{Related Work}


\textbf{Retrieval-Augmented Generation.} Retrieval-Augmented Generation (RAG)~\cite{rag-survey} has emerged as a powerful paradigm that incorporates external information or knowledge to enhance the quality, factuality, and relevance of generated text. 
Recent efforts~\cite{rag-borgeaud, Search-o1, WebThinker, Tool-Star} have leveraged RAG to address the challenge of hallucination and improve the performance of LLMs across a range of tasks. 
To furthur improve retrieval quality, several post-retrieval strategies~\cite{prca, replug, LongRefiner} have been introduced to fill the gap between retriever and generator, involving re-ranking, refinement and compression. 
Reranker~\citep{zhengbao_filter, bgm, reordering} reorder the retrieved results from retriever, enabling better alignment with the information needs of the LLM. 
Additionally, some studies~\cite{skr, self-rag} introduce techniques to mitigate noise in retrieved knowledge documents.
To tackle the problem of long-context limitation, various methods~\cite{RECOMP,fid-light} focus on compressing retrieved references to fit input length limits and removing irrelevant content to enhance robustness.

\textbf{Query Understanding.} Query Understanding~\cite{querysurveyazad, querysurveysong} encompasses a range of techniques aimed at improving the efficiency and accuracy of retrieval-augmented generation systems in the pre-retrieval stage, including query rewriting, disambiguation, decomposition and expansion. 
Recent advancements~\cite{queryllmanand, RQ-RAG} have highlighted the pivotal role of LLMs in query understanding in to enhance retrieval quality.
Query rewriting~\cite{query_rewrite, query_rewrite_min} involves reformulating the original query into a version more closely aligned with the information required for effective retrieval, thereby addressing the common mismatch between human intent and model interpretation.
Query disambiguation~\cite{dispeng, RaFe, ToC, EchoPrompt} focuses on clarifying user intent in ambiguous or multi-turn queries by transforming them into more specific and context-aware search inputs.
Query decomposition~\cite{Least-to-Most, Self-Ask, QCompiler} breaks down complex queries into simpler sub-queries to improve retrieval effectiveness and enable comprehensive answer generation. Reasoning-based query decomposition methods~\cite{ICAT, ReAct, plantimesrag,DBLP:conf/iclr/DongLLX0Z025} focus on generating reasoning traces or plans for solving complex tasks. 
Query expansion improves retrieval performance by enriching the original query with additional information derived from internal or external knowledge sources. Internal expansion methods~\cite{GenRead, Query2doc, HyDE, FLARE} focus on enhancing the original query using parametric knowledge within LLMs, while external expansion methods~\cite{LameR, GuideCQR} incorporate supplementary information from external sources such as knowledge bases.

\section{Method}
\textbf{Overview.} To ensure robust and high-quality RAG responses, we introduce the Omini-RAG pipeline, powered by LLMs, as shown in Figure~\ref{fig:main}. Given a query, an LLM first performs deep understanding, including rewriting and decomposition. Retrieval and reranking are then applied to obtain candidate documents for each sub-intent, followed by response generation via Falcon-10B. The subsequent subsections detail each stage of the pipeline.

\subsection{Problem Definition}
Compared to standard text generation, RAG often follows a \textit{retrieve-then-read} paradigm~\citep{lewis2021retrievalaugmented,DPA-RAG}, where an additional retriever is introduced to collect external knowledge and enhance the generation process.
Given the input user query be denoted by \(q\). The primary objective of the RAG system is to generate a comprehensive and relevant response \(R\). Formally, the system aims to find an optimal response \(R^*\) such that:
\begin{equation}
R^* = \arg\max_{R} P(R \mid q, \mathcal{K})
\end{equation}
where \(\mathcal{K}\) represents the available knowledge base or corpus from which information can be retrieved. The pipeline described below outlines the steps to approximate this optimal response.

\subsection{Query Understanding and Decomposition}

Real-world user queries often contain noise, such as spelling errors or ambiguous phrasing. Directly using such queries for searching can lead to inaccurate or irrelevant retrieval results. Therefore, we first employ an LLM for query understanding, rewriting. Let the original query be \(q\). The rewriting process can be represented as:
\begin{equation}
q' = \text{Rewrite}(q, \theta_{\text{rewrite}})
\end{equation}
where \(q'\) is the rewritten query, and \(\theta_{\text{rewrite}}\) represents the parameters of the LLM fine-tuned for the rewriting task.

After rewriting, although the semantics of the query \(q'\) become correct and intuitive, its intent may still be complex or multifaceted. To further achieve more precise retrieval targets and improve the recall of relevant documents, we perform query decomposition. In detail, the rewritten query \(q'\) is input to an LLM, which outputs a set of \(M\) sub-queries \(\{q'_{1}, q'_{2}, \dots, q'_{M}\}\). The above process can be formulated as:
\begin{equation}
\{q'_{s}\}_{s=1}^{M} = \text{Decompose}(q', \theta_{\text{decompose}})
\end{equation}
where each sub-query \(q'_{s}\) is designed to target a specific aspect of the original query. These sub-queries are typically generated in a structured format, such as JSON, for easy extraction and processing. This lays a solid foundation for retrieving broader and more accurate knowledge during the search process.

\subsection{Intent-Aware Knowledge Retrieval}

To obtain more comprehensive and extensive information, an intuitive approach is to retrieve information for each sub-intent derived from the decomposition of a complex query. For each decomposed sub-query \(q'_{s}\) (where \(s \in \{1, \dots, M\}\)), we utilize a search function to retrieve the top-K relevant documents from the knowledge base \(\mathcal{K}\). Let \(D_s\) be the set of documents retrieved for sub-query \(q'_{s}\):
\begin{equation}
D_s = \text{Search}(q'_{s}, \mathcal{K}, K) = \{d_{s,1}, d_{s,2}, \dots, d_{s,K}\}
\end{equation}
where \(d_{s,k}\) is the \(k\)-th document retrieved for sub-query \(q'_{s}\).
The initial set of retrieved documents, \(D_{\text{retrieved}}\), is then formed by taking the union of the documents retrieved for all sub-queries, \(D_{\text{retrieved}} = \bigcup_{s=1}^{M} D_s\). This ensures a broad coverage of information related to the different facets of the original query.

\subsection{Reranking and Generation:}

\textbf{Reranking:} 
To balance the redundant retrieval results introduced by sub-queries, reranking and filtering of the information are essential. After obtaining the initial set of retrieved documents \(D_{\text{retrieved}}\), we employ a reranking model to refine the selection and order of these documents. We use a sophisticated reranking model, such as BGE-reranker-large, for this purpose. The reranker computes a relevance score between the original query \(q\) (or the rewritten query \(q'\)) and each document \(d \in D_{\text{retrieved}}\).
Let \(\text{score}(q, d)\) be the relevance score assigned by the reranker. The documents in \(D_{\text{retrieved}}\) are then sorted based on these scores in descending order. We select the top-N documents from this sorted list to form the final set of context documents, \(D_{\text{reranked}}\):
\begin{equation}
D_{\text{reranked}} = \{d^*_1, d^*_2, \dots, d^*_N\} \subseteq D_{\text{retrieved}}
\end{equation}
such that \(\text{score}(q, d^*_i) \geq \text{score}(q, d^*_{i+1})\) for all \(i \in \{1, \dots, N-1\}\), and \(N\) is a predefined number of documents to be used for generation. After reranking, we obtained high-quality documents relevant to the query, providing essential support for accurate generation.

\noindent\textbf{Generation:} Finally, with the original query \(q\) and the top-N reranked documents \(D_{\text{reranked}}\), we use an LLM for generating the final response \(R\). The LLM is conditioned on both the query and the contextual information provided by the selected documents:
\begin{equation}
R = \text{Generate}(q, D_{\text{reranked}}, \theta_{\text{generate}})
\end{equation}
where \(\theta_{\text{generate}}\) represents the parameters of the LLM used for generation. This step aims to synthesize the information from the retrieved documents into a coherent, accurate, and contextually appropriate answer to the user's query.

\subsection{Pseudo Labeling and Evaluation}

Since the reference data does not contain ground-truth answers, we propose a pseudo-label generation and consistency evaluation strategy based on LLMs to support performance assessment and iterative optimization of RAG systems during development. It is important to note that the use of pseudo-labels strictly follows the competition guidelines: they are employed solely for system evaluation and analysis of answerless samples in dry tests, and are never used during the actual answer generation process.

Specifically, we first adopt Qwen2.5-7B-Instruct (fewer than 10B parameters)~\cite{qwen} as the reference model to generate pseudo answers by feeding it the input query along with its retrieved documents. To enable multi-dimensional evaluation, we define two core metrics—``relevance'' and ``faithfulness''—in accordance with the official evaluation criteria, and design dedicated prompts for each (below). 

Taking the relevance evaluation prompt as an example, it begins by defining the model's role, followed by key evaluation points and scoring guidelines. Additionally, four handcrafted examples are included to represent different rating levels, which enhance the model's contextual understanding and improve scoring consistency. The prompt for faithfulness evaluation follows the same structure as that of relevance. Finally, we employ Falcon-10B to independently execute both relevance and faithfulness evaluations, generating pseudo scores for each candidate answer accordingly.

\begin{tcolorbox}[
    colframe=gray,       
    colback=gray!5!white,             
    coltitle=white,                   
    coltext=black,                    
    fonttitle=\bfseries,              
    title=Relevance Evaluation Prompt Template,  
    boxrule=1pt,                      
    arc=2mm,                          
    width=\linewidth,                 
    left=7pt,                         
    right=7pt,                        
    top=5pt,                          
    bottom=5pt                        
]

\fontsize{8.5pt}{10pt}\selectfont
You are an expert evaluator assessing the quality of a predicted answer to a given question, using the provided reference answer (golden answer) as the standard. Your task is to assign a score based on the semantic equivalence and relevance of the prediction, based on the following scoring criteria:\\

Your evaluation considers:\\
- Equivalence: Does the Prediction convey the same meaning as the Golden Answer?\\
- Relevance: Does the Prediction directly address the Question without adding unrelated information? \\

Scoring Scale:\\
- 2: Correct and relevant (no irrelevant information).\\
- 1: Correct but contains irrelevant information.\\
- 0: No answer provided (abstention).\\
- -1: Incorrect answer.\\

Instructions:\\
Evaluate the prediction based on how well it aligns with the golden answer and how directly it addresses the question. Return only the numerical score (2, 1, 0, -1).\\

In-Context Examples:\{examples\}\\
Question: \{question\}\\
Golden Answer: \{answer\}\\
Prediction: \{prediction\}\\
Output:
\end{tcolorbox}

\begin{table}[t!]
    \centering
    \small
    \caption{Team Rankings of Session 1 of Live RAG Challenge}

    \begin{tabular}{clcc}
        \toprule
        \textbf{Rank} & \textbf{Team Name} & \textbf{Correctness} & \textbf{Faithful} \\
        \midrule
        1  & RMIT-ADMS            & 1.1993   & 0.4774   \\
        \textbf{2}  & \textbf{RUC\_DeepSearch (Ours)}      & \textbf{0.9693}   & \textbf{0.3878}   \\
        3  & Ped100X              & 0.9289   & 0.0434   \\
        4  & PRMAS-DRCA           & 0.9228   & 0.4106   \\
        5  & Hybrid Search w. Graph & 0.8751   & 0.3158   \\
        6  & BagBag               & 0.6941   & -0.9114  \\
        7  & UniClustRAG          & 0.6851   & 0.4601   \\
        8  & METURAG              & 0.6735   & 0.3253   \\
        9  & DeepRAG              & 0.5661   & 0.0978   \\
        10 & UiS-IAI              & 0.5523   & 0.4337   \\
        11 & SNU-LDILab           & 0.5174   & 0.1030   \\
        12 & Gravitational Lens   & 0.3766   & -0.9881  \\
        \bottomrule
    \end{tabular}
    \label{tab:team_rankings}
\end{table}

\begin{table}[t!]
\centering
\small

\caption{Performance comparison across different top-$k$ settings using OpenSearch. 5 (sc4) denotes generation using top-5 docs and 4 sampled reasoning paths for self-consistency.}
\begin{tabular}{lcccccccccc}
\toprule
\textbf{Method} & \textbf{top-$k$} & \multicolumn{5}{c}{\textbf{Relevance}} & \multicolumn{4}{c}{\textbf{Faithfulness}} \\
\cmidrule(lr){3-7} \cmidrule(lr){8-11}
& & Avg & -1 & 0 & 1 & 2 & Avg & -1 & 0 & 1 \\
\midrule
\multirow{5}{*}{Omni-RAG} 
           & 1 & 140 & 4 & 2 & 44 & 50 & 62 & 4 & 30 & 66 \\
           & 2 & 136 & 6 & 2 & 42 & 50 & 66 & 2 & 30 & 68 \\
           & 3 & 146 & 6 & 0 & 36 & 58 & 70 & 0 & 30 & 70 \\
           & 4 & 154 & 2 & 0 & 40 & 58 & 76 & 2 & 20 & 78 \\
           & 5 & 156 & 4 & 0 & 32 & 64 & 80 & 0 & 20 & 80 \\
\midrule
\multirow{2}{*}{Omni-RAG}          
           & 5 (sc4) & 170 & 2 & 0 & 24 & 74 & 72 & 0 & 28 & 72 \\
           & 5 (sc8) & 148 & 4 & 0 & 40 & 56 & 80 & 0 & 20 & 80 \\
\bottomrule
\end{tabular}
\vspace{-1em}
\label{tab:sc}
\end{table}

\subsection{Experiment}

\paragraph{\textbf{Experiment Setup.}} We strictly follow the requirements of the LiveRAG competition, using OpenSearch to retrieve from the Falcon corpus, BGE as the reranker, and Falcon-10B as the generator.

It is worth noting that in Table~\ref{tab:sc}, we experiment with the self-consistency(sc) strategy~\citep{wang2022self}, and the metrics are based on in-house pseudo-relevance and faithfulness scores generated by Qwen2.5-72B-Instruct.

\paragraph{\textbf{Main Result.}} The main results are presented in the primary table, where our Team ($RUC\_DeepSearch$) achieved an overall second place among the 12 participating teams in Session-1.

Notably, compared to the third-place team, Team Ped100X, our system achieved over a 4\% improvement in Correctness and approximately a 34\% improvement in Faithfulness. Similarly, compared to Team BagBag, which ranked fifth overall, our system outperformed them by around 9\% in Correctness and about 7\% in Faithfulness. These results clearly demonstrate the reliability and effectiveness of our RobustRAG framework.

\paragraph{\textbf{Dry Test Analysis.}} Dry-Test serves as a representative evaluation setting in our study. To further analyze our RAG performance, we select 50 samples from the Dry-Test set and use Qwen2.5-7B-instruct 's answer based on the top-5 documents as references to evaluate our model. The key findings are as follows.

1. \textbf{Performance scales with document count:} As the number of retrieved documents increases, the Omni-RAG model exhibits strong scalability in generation quality. Improvements are observed in both faithfulness and relevance, indicating that the performance benefits from richer document contexts.

2. \textbf{Trade-off in self-consistency path settings:} To enhance inference stability, we introduced a self-consistency mechanism. However, the 5 (sc8) configuration did not yield the expected performance gains, suggesting that more reasoning paths do not necessarily lead to better results. Interestingly, the 5 (sc4) setting improved relevance but led to a moderate decline in faithfulness, highlighting the need to balance path quantity with generation quality.

\section{Conclusion}
In this paper, we introduces Omni-RAG, a robust and scalable framework that enhances RAG systems through LLM-assisted query understanding. By integrating deep query decomposition, intent-aware retrieval, and reranking-guided generation, Omni-RAG effectively addresses the challenges of complex and noisy queries in live, open-domain environments. Our approach demonstrates strong potential for real-world applications, achieving Rank-2 overall performance in Session 1 of the SIGIR LiveRAG Challenge and offering a practical step toward more reliable and intelligent retrieval-augmented systems.
\balance
\bibliographystyle{ACM-Reference-Format}
\bibliography{sample-base}
\end{document}